\documentclass[10pt,twocolumn,letterpaper]{article}

\usepackage{wacv}              
\usepackage[accsupp]{axessibility}  

\usepackage{subcaption}
\usepackage{amssymb}
\usepackage{pifont}
\newcommand{\xmark}{\ding{55}}

\definecolor{wacvblue}{rgb}{0.21,0.49,0.74}
\usepackage[pagebackref,breaklinks,colorlinks,allcolors=wacvblue]{hyperref}
\usepackage[table]{xcolor}
\usepackage{multirow}

\def\wacvPaperID{1317} 
\def\confName{WACV}
\def\confYear{2026}

\title{SkelSplat: Robust Multi-view 3D Human Pose Estimation with Differentiable Gaussian Rendering} 

\author{Laura Bragagnolo\thanks{Corresponding author.}\\
University of Padova\\
{\tt\small bragagnolo@dei.unipd.it}
\and
Leonardo Barcellona\\
University of Amsterdam\\
{\tt\small l.barcellona@uva.nl}
\and
Stefano Ghidoni\\
University of Padova\\
{\tt\small ghidoni@dei.unipd.it}
}

\begin{document}
\maketitle

\begin{abstract}
Accurate 3D human pose estimation is fundamental for applications such as augmented reality and human-robot interaction. 
State-of-the-art multi-view methods learn to fuse predictions across views by training on large annotated datasets, leading to poor generalization when the test scenario differs.
To overcome these limitations, we propose SkelSplat, a novel framework for multi-view 3D human pose estimation based on differentiable Gaussian rendering. 
Human pose is modeled as a skeleton of 3D Gaussians, one per joint, optimized via differentiable rendering to enable seamless fusion of arbitrary camera views without 3D ground-truth supervision.
Since Gaussian Splatting was originally designed for dense scene reconstruction, we propose a novel one-hot encoding scheme that enables independent optimization of human joints.
SkelSplat outperforms approaches that do not rely on 3D ground truth in Human3.6M and CMU, while reducing the cross-dataset error up to 47.8\% compared to learning-based methods.
Experiments on Human3.6M-Occ and Occlusion-Person demonstrate robustness to occlusions, without scenario-specific fine-tuning.
Our project page is available here: \mbox{\url{https://skelsplat.github.io}}.

\end{abstract}
    
\section{Introduction}

\begin{figure}[t]
  \centering
  \includegraphics[width=0.75\columnwidth]{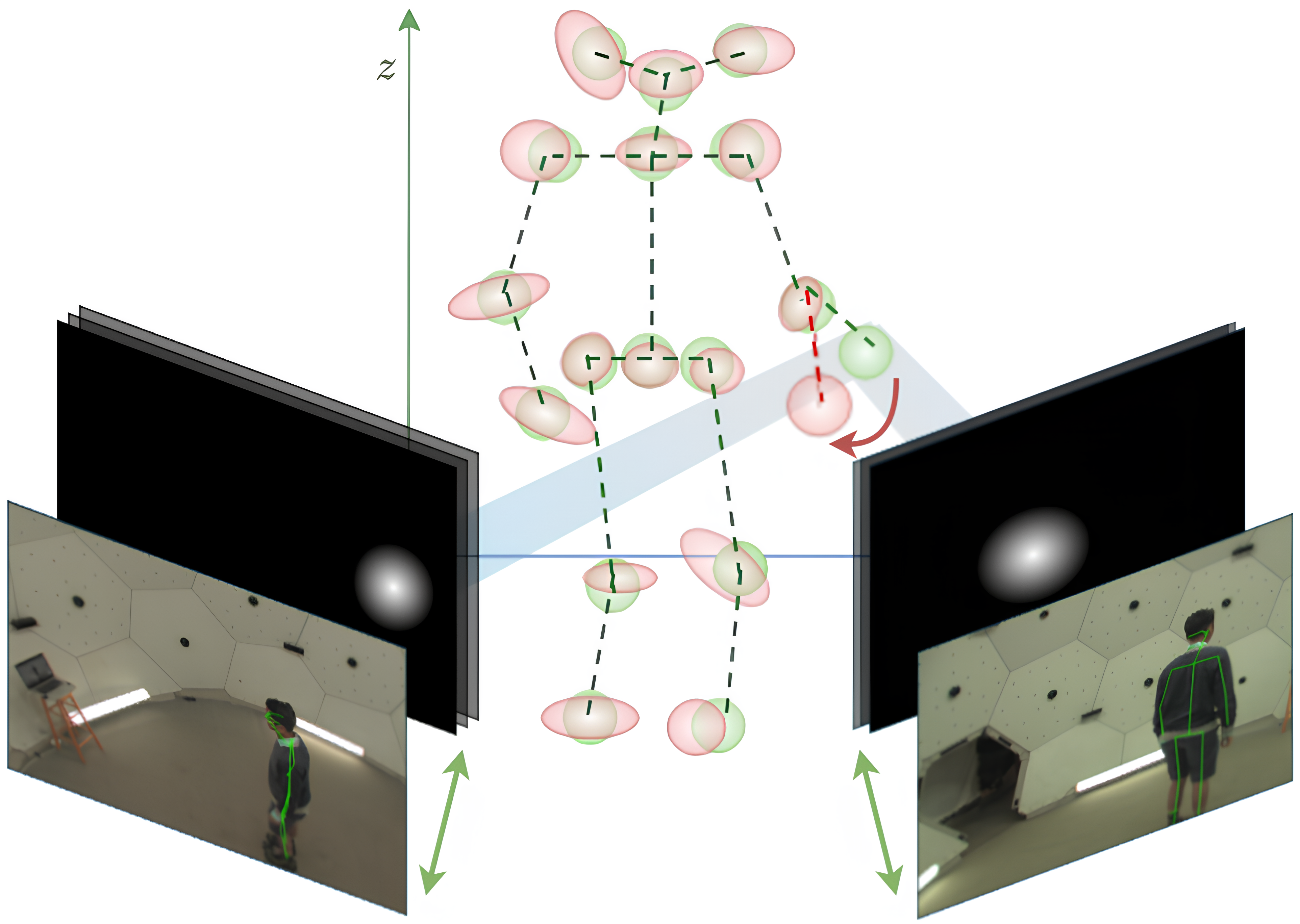}
  \caption{\textit{SkelSplat} 3D Gaussian joints are optimized (red) by aligning renderings with 2D detection heatmaps (green arrows).}
  \label{fig:framework-simple}
\end{figure}

Accurately estimating the three-dimensional human pose is a cornerstone for intelligent systems that perceive, interpret, and learn from human behavior. From autonomous driving~\cite{zheng2022multiAPP1} and human-robot collaboration~\cite{terreran2023generalAPP2} to animation~\cite{zhu2024champAPP3} and robot learning~\cite{zimmermann20183dAPP4}, applications across computer vision and AI increasingly demand robust 3D pose estimation. Among existing solutions, multi-view systems of calibrated cameras remain highly effective, as they triangulate 2D predictions to recover 3D poses~\cite{wan2023view, zhang2021adafuse, iskakov2019learnable, he2020epipolar}.
%
However, current state-of-the-art methods typically learn to fuse multi-view predictions directly from annotated datasets, making them tightly coupled to specific camera configurations~\cite{remelli2020lightweight}, pose distributions~\cite{moliner2024geometry}, and occlusion patterns~\cite{bragagnolo2025multi}. Consequently, performance often degrades sharply when these conditions change, requiring retraining or fine-tuning for each deployment scenario, reducing real-world applicability.\looseness=-1

In this work, we argue that multi-view fusion for 3D human pose estimation should be inherently flexible: it should generalize across arbitrary camera configurations and handle occlusions and appearance variations without retraining. To this end, we introduce \textit{SkelSplat}, a novel framework based on differentiable Gaussian rendering~\cite{kerbl20233d}.
%

%
%
SkelSplat leverages the Gaussian Splatting representation~\cite{kerbl20233d} to model each human joint as a 3D Gaussian. 
Given an initial 3D pose guess and 2D predictions from multiple cameras, \textit{SkelSplat} constructs a Gaussian representation by building an anisotropic Gaussian for each body joint and optimizes its position and shape by minimizing a rendering loss, as shown in~\cref{fig:framework-simple}.
%
%
%
%
We tailor the rendering function to human pose estimation by introducing a one-hot encoding scheme for joints. This allows our framework to robustly integrate multi-view cues without requiring scenario-specific training.

We evaluate \textit{SkelSplat} on standard benchmarks such as Human3.6M~\cite{ionescu2013human3} and CMU Panoptic Studio~\cite{joo2015panoptic}, showcasing its adaptability to diverse camera configurations. To assess robustness to occlusions, we further test on Human3.6M-Occ~\cite{bragagnolo2025multi}, where \textit{SkelSplat} achieves state-of-the-art performance, and on Occlusion-Person~\cite{zhang2021adafuse}, where it demonstrates strong accuracy without any training.
Notably, the proposed method removes the need for data collection and fine-tuning in new scenarios, thus facilitating scalable deployment.
%

In summary, our contributions are: 
\begin{enumerate}
\item We propose \textit{SkelSplat}, a novel framework for multi-view 3D human pose estimation, leveraging differentiable Gaussian rendering for view fusion;
\item We adapt Gaussian Splatting, primarily used for dense scene modeling, to skeleton-based 3D pose estimation;
\item We modify the original Gaussian Splatting rendering function to encode human joints using a one-hot representation, enabling pose-specific optimization;
\item We demonstrate that \textit{SkelSplat} achieves accurate 3D pose estimation under challenging occlusions and varying camera setups, without requiring retraining or fine-tuning.
\end{enumerate}

\begin{figure*}[ht]
  \centering
  \includegraphics[width=1.8\columnwidth]{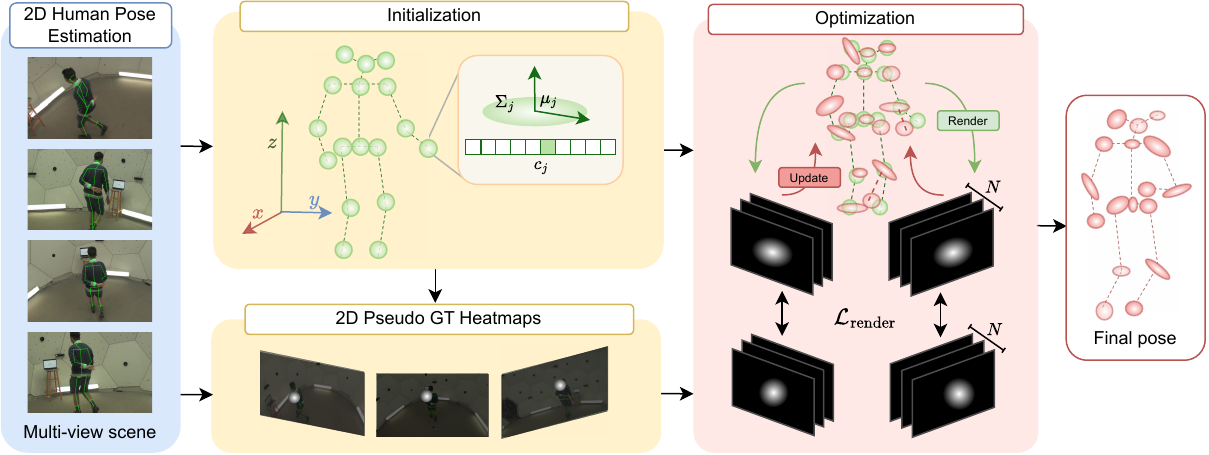}
  \caption{Overview of the \textit{SkelSplat} framework. Given multi-view images and 2D pose detections, we initialize a skeleton of 3D Gaussians, one per human joint. Pseudo ground truth heatmaps are generated from the 2D detections and used to supervise the optimization, which refines the Gaussians by minimizing a differentiable loss between heatmaps and Gaussian renderings.}
  \label{fig:framework}
\end{figure*}
\section{Related Work}


\paragraph{Gaussian Splatting}
Gaussian Splatting~\cite{kerbl20233d} had a disruptive impact on graphics~\cite{huang20242d, yu2024mip, guedon2024sugar, hollein20243dgs}, achieving state-of-the-art results in novel view synthesis. While subsequent works enhanced representations~\cite{huang20242d, yu2024mip}, reconstruction losses~\cite{guedon2024sugar}, or optimization techniques~\cite{hollein20243dgs}, our work focuses on applying the original formulation to multi-view 3D human pose estimation.
Gaussian Splatting has already been explored in diverse domains, including SLAM~\cite{matsuki2024gaussian}, real-to-sim transfer~\cite{jiang2025phystwin}, open-vocabulary segmentation~\cite{qin2024langsplat}, robot learning~\cite{barcellona2025dream}, and human reconstruction~\cite{kocabas2024hugs}, typically incorporating Gaussian Splatting without any fundamental changes to rendering or optimization pipelines. In contrast, we introduce a novel rendering function and supervision strategy, adapting Gaussian Splatting to the specific challenges of multi-view pose estimation.

\paragraph{Gaussian Splatting for 3D Human Reconstruction}
One of the closest applications of Gaussian Splatting for human perception is human reconstruction. Early works reconstruct accurate 3D human avatars from sparse images~\cite{kocabas2024hugs,Hu_2024_CVPR}, often incorporating parametric human models~\cite{Xiao_2025_CVPR, xiao2025rogsplat, svitov2024haha}, such as SMPL~\cite{bogo2016keep}, or skeletal priors~\cite{peng2023implicit, hu2024gauhuman, wen2024gomavatar} for articulated shapes and novel view generation.
Other approaches integrate diffusion for occlusion handling~\cite{sunoccfusion} or extract human features from Gaussian reconstructions~\cite{dey2024hfgaussian,prospero2025gst}, but primarily focus on high-quality offline reconstruction and subject-specific optimization. While these methods prioritize visual fidelity, our work instead focuses on accurate 3D joint localization.

\paragraph{Multi-view 3D Human Pose Estimation}
Multi-view human pose estimation aims to recover the 3D positions of each body joint by leveraging multiple synchronized cameras. Several works have explored learning-based fusion strategies to mitigate the reliance on accurate 2D detections, which often fail under occlusions~\cite{he2020epipolar, ma2021transfusion, zhang2021adafuse}. Iskakov et al.~\cite{iskakov2019learnable} aggregate 2D features into a shared 3D volume. 
He et al.~\cite{he2020epipolar} use epipolar geometry to attend to geometrically consistent pixels. TransFusion~\cite{ma2021transfusion} combines self-attention with positional encoding for occlusion robustness. AdaFuse~\cite{zhang2021adafuse} adaptively weights views to reduce the impact of low-quality detections.
While these methods handle occlusions, they remain limited in out-of-domain scenarios because their fusion strategies are learned from few collected datasets, such as Human3.6M~\cite{ionescu2013human3} and CMU Panoptic~\cite{joo2015panoptic}.

Recent works have therefore focused on robustness and generalization without relying on direct 3D supervision, using temporal consistency~\cite{davoodnia2024upose3d, moliner2024geometry}, geometric  consistency~\cite{zhao2023triangulation,bragagnolo2025multi}, multi-modal fusion~\cite{chen2024adaptivefusion} or optimization pipelines considering body priors learned on datasets~\cite{chen2022structural}.
While these approaches focus on learning fusion strategies, \textit{SkelSplat} employs differentiable Gaussian rendering to directly optimize 3D poses without relying on dataset-specific assumptions, demonstrating that a Gaussian representation, derived from Kerbl et al.~\cite{kerbl20233d}, generalizes better, especially in occluded scenes.

\paragraph{One-hot Encoding for Human Joints}
In many vision tasks, such as classification~\cite{he2016deep, dosovitskiy2020image} or semantic segmentation~\cite{chen2017deeplab, Cheng_2022_CVPR}, class identity is represented via one-hot encoding, with each class assigned to a dedicated output index or channel. 3D and 2D human pose estimation adopt the same principle, predicting joint coordinates~\cite{martinez2017simple,ye2023distilpose} or joint-specific heatmaps~\cite{newell2016stacked,xiao2018simple,sun2019deep}. 
This strategy has not yet been explored in Gaussian Splatting, where the rendered channels are typically RGB views of the reconstructed scene ~\cite{kerbl20233d}, not intended for accurate 3D points localization. 
To address this, we adapt Gaussian Splatting to operate on a one-hot encoding representation by redefining the rendering function and extending the output channels beyond RGB, enabling precise 3D joint localization.


\section{Method}

This section presents \textit{SkelSplat}, a novel method for multi-view 3D human pose estimation that leverages differentiable Gaussian rendering via Gaussian Splatting.
An overview of our framework is shown in~\cref{fig:framework}. 
In \textit{SkelSplat}, a skeleton $\mathrm{SK}=\{sk_0, \dots,sk_N\}$ is represented as a set of anisotropic 3D Gaussians $\mathrm{GS}=\{gs_0,\dots,gs_N\}$, in which each joint $sk_j$ is associated with a Gaussian $gs_j$.
The skeleton is optimized using the set of 2D keypoint detections $\mathrm{SK^{2D}_i}=\{sk^{2D}_{i0}, \dots, sk^{2D}_{iN}\}$ computed by a 2D human pose estimator on each camera $i$ of a set of $M$ synchronized and calibrated views, with $i \in \{1, \dots, M\}$. \textit{SkelSplat} uses the detections $\mathrm{SK^{2D}_i}$ to construct pseudo ground truth heatmaps $\mathrm{GS^{2D}_i}=\{gs_{i0}^{2D}, \dots,gs_{iN}^{2D}\}$ needed to optimize $\mathrm{GS}$ via differentiable rendering. This formulation allows for direct supervision from 2D keypoints alone, without requiring RGB supervision or 3D ground truth, making the method adaptable to varying camera configurations and robust in diverse scenarios.\looseness=-1

In the following, we first provide a brief overview of 3D Gaussian Splatting. 
We then present the details about the skeletal representation $\mathrm{GS}$, the pseudo ground truth $\mathrm{GS^{2D}_i}$, and the optimization process. 

\paragraph{Background}
Before delving into the details of \textit{SkelSplat}, we revise the classical definition of Gaussian Splatting~\cite{kerbl20233d}.
The objective of Gaussian Splatting is representing a scene as a set $\mathrm{G}$ of anisotropic Gaussians  to render novel views. Each Gaussian $g_i$ is defined by a point center ${p}_i = (x, y, z) \in \mathbb{R}^3$ and a covariance matrix $\Sigma_i \in \mathbb{R}^{3 \times 3}$, which together describe $g_i$'s spatial extent and orientation. The covariance is parameterized using a scale vector ${S}_i \in \mathbb{R}^3_+$ and a rotation matrix $R_i \in \mathrm{SO}(3)$, such that
$\Sigma_i = R_i S_i S_i^\top R_i^\top$.
Each Gaussian is also associated with an opacity $\alpha_i \in \mathbb{R}$ and a color representation ${c}_i$, modeled using spherical harmonics~\cite{ramamoorthi2001efficient} to support view-dependent radiance. During rendering, each 3D Gaussian is projected onto the 2D image plane using an affine approximation of the projective transformation. The projected 2D covariance $\Sigma_i^{2D}$ is computed as
$\Sigma_i^{2D} = J W \Sigma_i W^\top J^\top$,  
where $W$ is the view transformation matrix and $J$ is the Jacobian of the affine approximation.
In the original implementation, the set $\mathrm{G}$ is initialized from Structure-from-Motion (SfM), while the parameters of the Gaussians are optimized to minimize the photometric loss between the rendered images and the ground truth RGB images. During optimization, Gaussians densification and pruning are performed to better fit the entire scene.\looseness=-1

\paragraph{3D Human Pose as Skeleton of Gaussians}

Given an initial estimation of the 3D pose of a person $\mathrm{\hat{SK}} = \{\hat{sk}_0, \dots, \hat{sk}_N\}$, we represent the human pose as a skeleton of Gaussians $\mathrm{GS}$, in which each $gs_j = (\mu_j, \Sigma_j)$ represents $\hat{sk}_j$, where $\mu_j \in \mathbb{R}^3$ is the 3D position of $\hat{sk}_j$, and $\Sigma_j \in \mathbb{R}^{3 \times 3}$ is the covariance, empirically initialized to $\Sigma_j = 3 \cdot \mathbf{I}_3$, with $\mathbf{I}_3$ denoting the $3 \times 3$ identity matrix. The approach is agnostic to the initialization method; therefore $\mathrm{GS}$ can be constructed by different approaches, such as triangulation or monocular pose estimation~\cite{bragagnolo2025multi}.\looseness=-1



\paragraph{Joint encoding and rendering} 
During the optimization process, we need to ensure that each $gs_j$ is supervised according to the correct joint prediction $sk^{2D}_{ij}$. Thus, we modify the RGB-based appearance encoding $c_j$ with a joint identity encoding that activates only the $j$-th channel:
\begin{equation}
{c}_j[k] = \begin{cases}
1 & \text{if } k = j \\
0 & \text{otherwise}
\end{cases}
\quad \text{for } k = 1, \dots, N.
\end{equation}
This identity vector is encoded using degree-zero spherical harmonics coefficients, repurposing the appearance field to represent joint identity rather than color. However, this change would not be effective without modifying the rendering function that splats the Gaussians onto the image plane. 
We modify this function to produce a $N$-channel tensor, where each channel corresponds to the splat of a single joint Gaussian $gs_j$ on a camera view. This allows independent supervision of $gs_j$, even when multiple 2D joint detections $sk^{2D}_{ij}$ are overlapping in the image plane.\looseness=-1


\paragraph{Supervision from 2D Detections}


To supervise the optimization, we exploit geometric cues only, to enable flexible supervision across diverse datasets. To do this, we generate pseudo ground truth heatmaps from 2D keypoint detections obtained by a pre-trained model.\looseness=-1

For each camera view $i \in \{1, \dots, M\}$, we generate a set of $N$ pseudo ground truth images $\{I_{ij}\}_{j=1}^{N}$, one for each joint. Each image $I_{ij} \in \mathbb{R}^{H \times W}$ is constructed by rendering a 2D Gaussian $gs_{ij}^{2D}$ centered at the detected 2D keypoint ${sk}_{ij}^{2D} \in \mathbb{R}^2$, with a covariance matrix $\Sigma_{ij}^{2D} \in \mathbb{R}^{2 \times 2}$. Covariance in 2D joint heatmaps must be geometrically consistent with the 3D Gaussian representation: 3D joints closer to the camera should exhibit larger projected covariances due to perspective scaling, while distant joints should appear smaller.
To ensure multi-view geometric consistency, we compute $\Sigma^{2D}_{ij}$, 2D covariance of joint $j$ in view $i$, by reprojecting the 3D covariance $\Sigma_j$ of the initial 3D pose estimate:
\begin{equation}
\Sigma^{2D}_{ij} = J_i W_i \Sigma_j W^T_i J^T_i \,,
\label{eq:cov-reprojection}
\end{equation}
where ${W}_i \in \mathbb{R}^{4 \times 4}$ is the camera extrinsic transformation (world-to-camera), and ${J}_i \in \mathbb{R}^{2 \times 3}$ is the Jacobian matrix of the perspective projection at the joint position in camera coordinates.\looseness=-1


This construction yields 2D joint heatmaps whose Gaussian shape accurately reflects the 3D joint Gaussians under perspective projection, enabling precise and geometrically consistent multi-view optimization.\looseness=-1

\paragraph{Optimization}

The human pose, represented using $\mathrm{GS}$, is splatted across all camera views and refined to minimize the masked $\mathcal{L}_2$ loss between the pseudo ground truth images and the rendered images:
%
\begin{equation}
\mathcal{L}_{\text{render}} =
\sum_{j=1}^N \sum_{i=1}^M
\left\| \mathcal{M}_{ij} \odot \left( I^{\text{render}}_{ij} - I^{\text{pseudo}}_{ij} \right) \right\|_2^2 \,.
\end{equation}
Here, $N$ is the number of joints, $M$ the number of camera views, and $I^{\text{render}}_{ij}$ and $I^{\text{pseudo}}_{ij}$ are the rendered and pseudo ground truth heatmaps for joint $j$ in camera $i$, with the latter generated according to~\cref{eq:cov-reprojection}. $\mathcal{M}_{ij}$ is a binary mask that selects only pixels with non-zero values in either $I^{\text{render}}_{ij}$ or $I^{\text{pseudo}}_{ij}$.
The choice of a masked $\mathcal{L}_2$ loss over the standard one is due to the high presence of background regions that can reduce convergence. \looseness=-1

We introduce a 3D structural symmetry loss that regularizes the lengths of symmetric limbs, such as arms or legs, to encourage the recovered 3D structure to be anatomically coherent even in the presence of noisy observations, such as in the case of occlusions. For each pair of symmetric limbs, we penalize inconsistencies in their lengths:
\begin{equation}
\mathcal{L}_{\text{sym}} = \sum_{(l, r) \in \mathcal{S}} \left( \left\| \mathbf{p}_l^{1} - \mathbf{p}_l^{2} \right\|_2 - \left\| \mathbf{p}_r^{1} - \mathbf{p}_r^{2} \right\|_2 \right)^2 \,,
\end{equation}
where $\mathcal{S}$ denotes the set of symmetric limb pairs; ${p}_l^{1}$ and ${p}_l^{2}$ are the 3D endpoints of the left limb, and ${p}_r^{1}$ and ${p}_r^{2}$ are the endpoints of the corresponding right limb. This term acts as a regularizer that complements the view-based loss by promoting globally consistent pose estimations.\looseness=-1

The total loss used during optimization is a combination of the masked multi-view $\mathcal{L}_2$ loss and the 3D structural loss:
\begin{equation}
\mathcal{L}_{\text{opt}} = \mathcal{L}_{\text{render}} + \lambda_{\text{sym}} \mathcal{L}_{\text{sym}} \,,
\end{equation}
where $\lambda_{\text{sym}}$ is a weighting factor that balances the influence of the symmetry term, empirically set to 1e-5.\looseness=-1

Unlike standard Gaussian Splatting, which updates the 3D Gaussians independently for each view after computing the loss, we accumulate gradients across all views, improving stability and cross-view coherence. In this way, cues from views are more effectively merged, allowing the optimization to better handle partial or occluded observations from individual views that might bias the optimization. Gaussian pruning and densification are removed to avoid changing the cardinality of $\mathrm{GS}$; 
the optimization, based on Adam, is run for up to 125 iterations, while early stopping is triggered when the difference between the minimum loss in two consecutive windows of size $M$ is less than 1e-6.\looseness=-1

\begin{figure}[t]
  \centering
  \includegraphics[width=0.8\columnwidth]{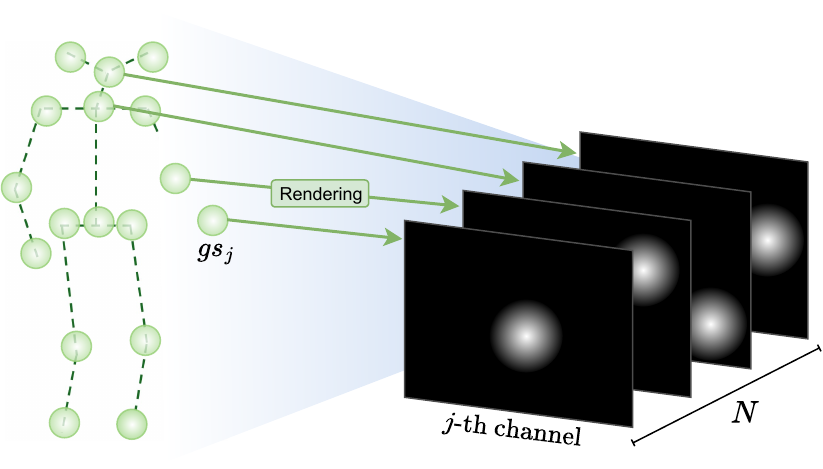}
  \caption{One-hot encoding, each joint rendered to its channel.\looseness=-1}
  \label{fig:encoding}
\end{figure}

\section{Experiments}
We evaluate our approach on four multi-view datasets to assess its robustness under different levels of occlusion and to test generalization to novel scenarios.

\subsection{Datasets}
\paragraph{Human3.6M} Human3.6M~\cite{ionescu2013human3} is a widely used benchmark for 3D human pose estimation. It contains 3.6 million frames recorded from four synchronized cameras.
The dataset includes 11 subjects performing everyday activities, such as walking and eating.
Following standard protocol, subjects S9 and S11 are used for evaluation.

\paragraph{CMU Panoptic Studio} The CMU Panoptic Studio~\cite{joo2015panoptic} provides rich multi-view recordings of people engaged in various activities and social interactions. Its capture setup includes 480 VGA cameras, 31 HD cameras, and 10 Kinect sensors. In this study, we use only sequences featuring a single person, following Xiang et al.~\cite{Xiang_2019_CVPR}.

\paragraph{Human3.6M-Occ}
To evaluate our approach on challenging occlusion scenarios, we consider Human3.6M-Occ~\cite{bragagnolo2025multi}, a modified version of Human3.6M that introduces occlusions by overlaying objects and animals from the Pascal VOC 2012 dataset~\cite{pascal-voc-2012} on the original images. We consider three configurations:
\textit{Human3.6M-Occ-2}, with 2 views occluded out of 4,
\textit{Human3.6M-Occ-3}, with 3 views occluded,
and \textit{Human3.6M-Occ-3-Hard}, which features larger occluders over the subject's body.

\paragraph{Occlusion-Person}
Occlusion-Person~\cite{zhang2021adafuse} is a synthetic dataset comprising 13 human models animated with poses from the CMU Motion Capture database and rendered in 9 indoor scenes. Heavy occlusions are introduced using common objects like sofas and desks. Each scene is captured from 8 evenly spaced camera views arranged in a circle.

\paragraph{Evaluation}
We assess 3D pose estimation accuracy using the Mean Per Joint Position Error (MPJPE), which measures the average Euclidean distance in millimeters between predicted and ground-truth 3D joint coordinates.
We take as the predicted 3D joint positions the means of the 3D Gaussians produced by \textit{SkelSplat}. 
Unlike prior works, such as~\cite{zhang2021adafuse}, that often report root-relative MPJPE, we consider absolute MPJPE, assessing both joint configuration and global position accuracy without aligning predicted and ground-truth poses at the root joint, since we are interested in the absolute pose in the 3D space.
Lower MPJPE values indicate more precise 3D joint predictions.

\subsection{Results}

\paragraph{Results on H36M}
We evaluate \textit{SkelSplat} on the Human3.6M dataset to compare against state-of-the-art methods, as shown in~\cref{tab:h36m}.
We evaluate three variants of \textit{SkelSplat}, each using different source of 2D keypoints: MeTRAbs~\cite{Sarandi_2023_WACV}, CPN~\cite{chen2018cascaded} (as used in~\cite{pavllo20193d}), and ResNet-152 (as in~\cite{zhang2021adafuse}), with the latter two fine-tuned on Human3.6M.
Using accurate, dataset-specific 2D keypoints ensures fair comparison with prior works, which fine-tune 2D pose estimators on the dataset of interest ~\cite{davoodnia2024upose3d, zhao2023triangulation, iskakov2019learnable, ma2021transfusion}, as body joints definitions differ across datasets~\cite{Sarandi_2023_WACV}.
MeTRAbs, trained on a large variety of datasets, is used off-the-shelf.
While using 2D detected poses for optimization can indeed limit quality, this affects all multi-view methods, as they depend on 2D poses for fusion~\cite{ma2021transfusion}. 
For the initial 3D pose estimate, we adopt the method proposed in~\cite{bragagnolo2025multi}, which combines monocular 3D predictions by minimizing 2D reprojection error. An ablation comparing initialization strategies is provided in~\cref{sec:ablations}.
Among the variants, the best performance is achieved using 2D keypoints from ResNet-152, which outperforms strong baselines such as~\cite{moliner2024geometry, davoodnia2024upose3d, zhao2023triangulation, zhang2021adafuse, ma2021transfusion}.
Our method is surpassed only by Isakov et al.~\cite{iskakov2019learnable} which, however, relies on full 3D supervision and extensive training on Human3.6M. In contrast, \textit{SkelSplat} does not require any 3D ground truth during training, optimizing multi-view scenes using only detected 2D keypoints, making it significantly more adaptable.\looseness=-1

\paragraph{Cross dataset generalization} We evaluate cross-dataset generalization in~\cref{tab:generalization-h36m-cmu}, where we compare our method to~\cite{iskakov2019learnable, moliner2024geometry, bartol2022generalizable} trained on the CMU Panoptic dataset and tested on Human3.6M. 
Results clearly show that, while learning-based approaches retain excellent performance in the training domain, they significantly drop when applied to different datasets.\looseness=-1

\begin{table*}[tb]
\centering
\resizebox{2\columnwidth}{!}{%
\begin{tabular}{@{}l|c|ccccccccccccccc|c@{}}
\toprule
Absolute MPJPE, mm                                                      & 3D GT      & Dir.          & Disc.         & Eat           & Greet         & Phone         & Photo         & Pose          & Purch.        & Sit           & SitD.         & Smoke         & Wait          & WalkD.        & Walk          & WalkT.        & Avg $\downarrow$ \\ \midrule \midrule
Tome et al.~\cite{tome2018rethinking}                                   & \checkmark & 43.3          & 49.6          & 42.0          & 48.8          & 51.1          & 64.3          & 40.3          & 43.3          & 66.0          & 95.2          & 50.2          & 52.2          & 51.1          & 43.9          & 45.3          & 52.8             \\
Cross-view Fusion~\cite{qiu2019cross}                                   & \checkmark & 28.9          & 32.5          & 26.6          & 28.1          & 28.3          & 29.3          & 28.0          & 36.8          & 41.0          & 30.5          & 35.6          & 30.0          & 28.3          & 30.0          & 30.5          & 31.2             \\
Remelli et al.~\cite{remelli2020lightweight}                            & \checkmark & 27.3          & 32.1          & 25.0          & 26.5          & 29.3          & 35.4          & 28.8          & 31.6          & 36.4          & 31.7          & 31.2          & 29.9          & 26.9          & 33.7          & 30.4          & 30.2             \\
Generalizable Triang.~\cite{bartol2022generalizable}                    & \checkmark & 27.5          & 28.4          & 29.3          & 27.5          & 30.1          & 28.1          & 27.9          & 30.8          & 32.9          & 32.5          & 30.8          & 29.4          & 28.5          & 30.5          & 30.1          & 29.1             \\
Epipolar Transformer~\cite{he2020epipolar}                              & \checkmark & 25.7          & 27.7          & 23.7          & 24.8          & 26.9          & 31.4          & 24.9          & 26.5          & 28.8          & 31.7          & 28.2          & 26.4          & 23.6          & 28.3          & 23.5          & 26.9             \\
TransFusion~\cite{ma2021transfusion}                                    & \checkmark & 24.4          & 26.4          & 23.4          & 21.1          & 25.2          & 23.2          & 24.7          & 33.8          & 29.8          & 26.4          & 26.8          & 24.2          & 23.2          & 26.1          & 23.3          & 25.8             \\
AdaFuse~\cite{zhang2021adafuse}                                         & \checkmark & -             & -             & -             & -             & -             & -             & -             & -             & -             & -             & -             & -             & -             & -             & -             & 24.3             \\
Geometry Transformer~\cite{moliner2024geometry}                         & \checkmark & -             & -             & -             & -             & -             & -             & -             & -             & -             & -             & -             & -             & -             & -             & -             & 22.7             \\
Iskakov et al. - Vol~\cite{iskakov2019learnable}                        & \checkmark & \textbf{18.0} & \textbf{18.3} & \textbf{16.5} & \textbf{16.1} & \textbf{17.4} & \textbf{18.2} & \textbf{16.5} & \textbf{18.5} & \textbf{19.4} & \textbf{20.1} & \textbf{18.2} & \textbf{17.4} & \textbf{17.2} & \textbf{19.2} & \textbf{16.6} & \textbf{17.7}    \\ \midrule \midrule
UPose3D~\cite{davoodnia2024upose3d}                                     & \xmark     & -             & -             & -             & -             & -             & -             & -             & -             & -             & -             & -             & -             & -             & -             & -             & 26.2             \\
TRL~\cite{zhao2023triangulation}                                        & \xmark     & -             & -             & -             & -             & -             & -             & -             & -             & -             & -             & -             & -             & -             & -             & -             & 25.8             \\
MV Pose Fusion~\cite{bragagnolo2025multi}                               & \xmark     & 23.8          & 24.6          & 23.4          & 24.1          & 26.9          & 24.0          & 25.3          & 28.3          & 31.7          & 25.9          & 26.9          & 24.2          & 23.0          & 27.3          & 24.2          & 25.6             \\
Iskakov et al. - Alg~\cite{iskakov2019learnable}                        & \xmark     & 21.7          & 23.7          & 22.2          & 20.4          & 26.7          & 24.2          & 19.9          & 22.6          & 31.2          & 35.6          & 26.8          & 21.2          & 20.9          & 24.6          & 21.1          & 24.5             \\
RANSAC (as in~\cite{iskakov2019learnable})                              & \xmark     & 21.6          & 22.9          & 20.9          & 21.0          & 23.1          & 23.0          & 20.8          & 22.0          & 26.4          & 26.6          & 24.0          & 21.5          & 21.0          & 23.9          & 20.8          & 22.8             \\ \hline
\rowcolor{blue!8} \textbf{SkelSplat} (MetrAbs~\cite{Sarandi_2023_WACV}) & \xmark     & 24.4          & 25.9          & 24.7          & 25.0          & 28.2          & 24.4          & 25.1          & 28.2          & 30.3          & 26.9          & 28.7          & 25.0          & 26.9          & 28.1          & 28.7          & 26.7             \\
\rowcolor{blue!8} \textbf{SkelSplat} (CPN)                              & \xmark     & 25.7          & 25.4          & 25.0          & 24.6          & 27.1          & 23.4          & 23.7          & 27.1          & 28.7          & 27.1          & 28.5          & 24.3          & 26.3          & 28.4          & 26.9          & 26.2             \\
\rowcolor{blue!8} \textbf{SkelSplat} (ResNet-152)                       & \xmark     & \textbf{18.9}          & \textbf{20.1}         & \textbf{19.5}          & \textbf{18.1}          & \textbf{20.6}          & \textbf{18.6}         & \textbf{20.2}          & \textbf{22.2}          & \textbf{23.5}          & \textbf{21.6}          & \textbf{20.8}          & \textbf{18.8}          & \textbf{19.1}          & \textbf{22.2}          & \textbf{18.6}          & \textbf{20.3}             \\ \bottomrule \bottomrule
\end{tabular}%
}
\caption{Overall comparison with state-of-the-art methods on Human3.6M, where `\checkmark' indicates training with 3D GT.}
\label{tab:h36m}
\end{table*}

\begin{table}[tb]
\centering
\resizebox{\columnwidth}{!}{%
\begin{tabular}{lcccc}
\toprule
Absolute MPJPE, mm                                                      & Train                   & Test                     & Avg $\downarrow$ & Impr. (\%) $\uparrow$ \\ \midrule \midrule
Geometry-biased transformer~\cite{moliner2024geometry}                  & CMU                     & H36M                     & 38.9             & -                     \\
Iskakov et al. - Vol~\cite{iskakov2019learnable}                        & CMU                     & H36M                     & 34.0             & + 12.6                \\
Generalizable Triang.~\cite{bartol2022generalizable}                    & \multicolumn{1}{l}{CMU} & \multicolumn{1}{l}{H36M} & 31.0             & + 20.3                \\
\rowcolor{blue!8} \textbf{SkelSplat} (MeTRAbs~\cite{Sarandi_2023_WACV}) & -                       & H36M                     & \underline{26.0}   & \underline{+ 31.4}      \\
\rowcolor{blue!8} \textbf{SkelSplat} (ResNet-152)                       & -                       & H36M                     & \textbf{20.3}    & \textbf{+ 47.8}       \\ \bottomrule \bottomrule
\end{tabular}%
}
\caption{Comparison between models trained on CMU and evaluated on H36M and \textit{SkelSplat} tested on H36M without any training.}
\label{tab:generalization-h36m-cmu}
\end{table}


\begin{table}[tb]
\centering
\resizebox{0.6\columnwidth}{!}{%
\begin{tabular}{lc}
\toprule 
Absolute MPJPE, mm                                                          &    Avg $\downarrow$  \\ \midrule \midrule
RANSAC (as in~\cite{iskakov2019learnable})                                   & 39.5 \\
Voxelpose~\cite{tu2020voxelpose}                                                & 25.5 \\
Generalizable Triangulation~\cite{bartol2022generalizable}                       & 25.4 \\
Iskakov et al. - Alg~\cite{iskakov2019learnable}                             & \underline{21.3} \\
\rowcolor{blue!8} \textbf{SkelSplat} (MeTRAbs~\cite{Sarandi_2023_WACV}) &\textbf{ 20.9}  \\ \bottomrule \bottomrule
\end{tabular}%
}
\caption{Results on the CMU Panoptic Studio dataset, considering single person sequences in a 4-camera setup.}
\label{tab:cmu}
\end{table}

\paragraph{Results on CMU Panoptic Studio} On this dataset we compare against baselines designed for cross-scene generalization, such as~\cite{bartol2022generalizable, tu2020voxelpose}, and with Algebraic Triangulation and RANSAC from~\cite{iskakov2019learnable}. 
We consider a 4-views setup, selecting cameras 1, 2, 10 and 13, following the configuration proposed in~\cite{iskakov2019learnable}.
In~\cref{tab:cmu} we report results obtained using 2D poses from MeTRAbs and considering the triangulation of 2D detections as initial guess.
\textit{SkelSplat} achieves an absolute MPJPE of 20.9\,mm, outperforming other approaches aimed at cross-scene generalization. Notably, it surpasses both RANSAC and Algebraic Triangulation even when those methods use fine-tuned 2D poses: while these methods struggle with camera configurations that produce significant self-occlusions, \textit{SkelSplat} shows a better ability in handling challenging visual conditions.\looseness=-1

\begin{figure}[bh]
  \centering
  \includegraphics[width=\columnwidth]{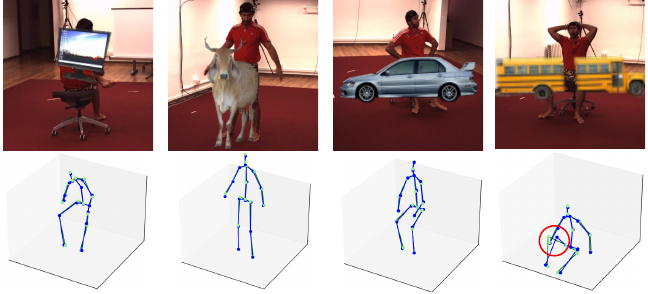}
  \caption{\textit{SkelSplat} results (blue) on Human3.6M-Occ-3 with ground-truth poses (green). The rightmost column shows a failure case under occlusion, where the left knee is incorrectly predicted.}
  \label{fig:results}
\end{figure}

\paragraph{Results on Human3.6M-Occ} We assess the robustness of \textit{SkelSplat} to occlusions using the Human3.6M-Occ dataset, which introduces synthetic occlusions of varying severity. For joints that are more prone to being occluded, such as elbows, wrists, knees, and ankles, we slightly enlarge their initial covariance by applying a scaling factor of 1.25. This provides more flexibility during optimization, enabling the model to better cope with inaccurate 2D detections and to more effectively explore plausible joint configurations under occlusion. The effect of this covariance scaling is further analyzed in~\cref{sec:ablations}. We initialize our 3D pose with monocular predictions fusion as in~\cite{bragagnolo2025multi}.
As shown in~\cref{tab:h36m-occ}, \textit{SkelSplat} achieves state-of-the-art performance across all three Human3.6M-Occ benchmarks, demonstrating superior robustness compared to view-fusion methods specifically designed to handle occlusions~\cite{ma2021transfusion, zhang2021adafuse, bragagnolo2025multi}. The best results are obtained using 2D poses from a ResNet-152 model trained solely on the original Human3.6M dataset—the same model used in AdaFuse~\cite{zhang2021adafuse}. In the most challenging Human3.6M-Occ-3-Hard setting, \textit{SkelSplat} ranks second only to~\cite{bragagnolo2025multi}, which benefits from additional pose confidence weighting during fusion. In contrast, \textit{SkelSplat} achieves competitive results without using any confidence scores or learned weights, relying only on the optimization using the 2D pose detections. Compared to standard non-learning baselines such as triangulation and RANSAC, which suffer significant performance degradation under heavy occlusion, \textit{SkelSplat} maintains a high accuracy, showing robustness to incomplete and noisy 2D observations. Some qualitative results are shown in~\cref{fig:results}.\looseness=-1

\paragraph{Results on Occlusion-Person} We further evaluate the occlusion robustness of \textit{SkelSplat} on the Occlusion-Person dataset, including baseline methods such as Algebraic Triangulation, RANSAC, and AdaFuse~\cite{zhang2021adafuse}. For this experiment, we report results using both 8-camera and 4-camera configurations, selecting cameras 1, 3, 5, and 7 for the latter. As initial pose, we use the triangulation of 2D keypoints predicted by a ResNet-50 model, following the same setup as AdaFuse~\cite{zhang2021adafuse}.
SkelSplat outperforms all baselines that were not trained on the target dataset, improving upon the best-performing model by 5.3\,mm and 5.6\,mm, on the 4-view and 8-view settings, respectively. 
Note that for AdaFuse we report two variants: one trained on Human3.6M and one on Occlusion-Person. 
While the latter achieves better performance than \textit{SkelSplat}, this comparison is less meaningful, as its performance significantly degrades when applied outside its training domain.
 
Overall, these experiments highlight \textit{SkelSplat}’s ability to generalize across domains, camera configurations, and occluded conditions while maintaining high accuracy, making it easily applicable to novel environments.\looseness=-1 

\begin{table}[t]
\centering
\resizebox{\columnwidth}{!}{%
\begin{tabular}{lccc}
\toprule
Absolute MPJPE, mm                                          & H3.6M-Occ-2 & H3.6M-Occ-3 & H3.6M-Occ-3h \\ \midrule \midrule
Alg. Triangulation (ResNet-152)                             & 43.2        & 48.9        & 120.4          \\
TransFusion~\cite{ma2021transfusion}                       & 40.8        & 76.3            & 96.5       \\
RANSAC (as in~\cite{zhang2021adafuse})                      & 33.7        & 38.6        & 80.7           \\
Alg. Triangulation (MeTRAbs~\cite{Sarandi_2023_WACV})                                & 36.0        & 39.0        &  67.5         \\ 
AdaFuse~\cite{zhang2021adafuse}                              & 27.9        & 31.2        & 41.1           \\
MV Pose Fusion~\cite{bragagnolo2025multi}                & 33.4        & 36.7        & \underline{37.8}           \\
\rowcolor{blue!8} \textbf{SkelSplat} (MeTRAbs~\cite{Sarandi_2023_WACV})        & \underline{29.6}        & \underline{31.1}        &   38.1       \\
\rowcolor{blue!8} \textbf{SkelSplat} (ResNet-152)     & \textbf{24.6}        & \textbf{27.0}        &   \textbf{34.8}       \\ \bottomrule \bottomrule
\end{tabular}%
}
\caption{Results on the three benchmarks of the Human3.6M-Occ dataset, which introduce occlusions on 2 views (Occ-2), 3 views (Occ-3), and severe occlusions on 3 views (Occ-3-Hard).}
\label{tab:h36m-occ}
\end{table}

\begin{table}[tb]
\centering
\resizebox{0.6\columnwidth}{!}{%
\begin{tabular}{lcc}
\toprule
Absolute MPJPE, mm                                       & 4 views           & 8 views \\ \midrule \midrule
Alg. Triangulation                                        & 59.1              & 49.2    \\
RANSAC (as in~\cite{zhang2021adafuse})                   & 71.4               & 39.2    \\
Adafuse~\cite{zhang2021adafuse}\dag                     & \textbf{19.5}      & \textbf{12.6}    \\
Adafuse~\cite{zhang2021adafuse}                          & 45.6                 & 36.1    \\
\rowcolor{blue!8} \textbf{SkelSplat} (ResNet-50)         & \underline{40.3}    & \underline{30.5}    \\ \bottomrule \bottomrule
\end{tabular}%
}
\caption{Results on the Occlusion-Person dataset, for 8-camera and 4-camera setups. `\dag' indicates training on the target dataset.}
\label{tab:occ-pers}
\end{table}

\subsection{Ablation Studies}
\label{sec:ablations}

To understand the contribution of each component, we conduct ablation studies analyzing the impact of our one-hot encoding, optimization design choices such as cross-view gradient accumulation and the 3D loss term, input conditions including camera selection, initialization strategies, and image resolution, as well as covariance scaling in pseudo ground-truth generation.\looseness=-1

\paragraph{Impact of one-hot joint encoding} 
On Human3.6M, we evaluate our one-hot encoding for joint Gaussians. We compare it against the standard RGB formulation of~\cite{kerbl20233d}, which uses three shared channels, and a single-channel variant where all Gaussians are aggregated. As reported in~\cref{tab:abl-encoding}, our one-hot encoding reduces absolute MPJPE by up to 10\% relative to these alternatives.\looseness=-1

\paragraph{Impact of different cameras} We analyze the effect of the number of input views on \textit{SkelSplat} on the CMU Panoptic Studio dataset. In~\cref{tab:abl-views}, we report results for varying camera configurations using 4, 5, 6, 7, and 8 views. The 4-camera setup includes views 1, 2, 10, and 13, with additional views progressively added: camera 3 (5 views), camera 23 (6 views), camera 19 (7 views), and camera 30 (8 views).
The approach clearly improves by increasing the number of views. We observe a consistent reduction in error as more cameras are added, with a slight deviation in the 7-view setup due to oblique angle and subject cropping in camera 19. Overall, this trend suggests that \textit{SkelSplat} performance improves in dense camera scenarios. 
\looseness=-1

\paragraph{Contribution of cross-view gradient accumulation}
On Human3.6M, we evaluate the impact of accumulating optimization gradients across multiple views. As shown in~\cref{tab:abl-accumulation}, we compare standard single-view gradient updates (as in the original Gaussian Splatting) with gradient accumulation over 2 views and 4 views, with the latter yielding the best performance. Notably, even accumulating over just 2 views results in a 2.7\,mm improvement over the standard approach, highlighting that incorporating information from multiple views before parameter updates promotes optimization stability and cross-view consistency.\looseness=-1

\paragraph{Impact of initialization and noise}
On Human3.6M, we analyze how initialization affects 3D joint optimization. In~\cref{tab:abl-initial-guess}, we compare \textit{SkelSplat} initialized with simple algebraic triangulation and with multi-view fusion of 3D monocular poses from an off-the-shelf estimator~\cite{bragagnolo2025multi}, using 2D detections from MeTRAbs~\cite{Sarandi_2023_WACV}, CPN and ResNet-152. 
Results show that the initialization strategy has little influence on the final accuracy, as different starting points converge to comparable MPJPE values. 
To further assess robustness, we perturb triangulated initializations with Gaussian noise of increasing standard deviation applied independently to each joint, as reported in~\cref{tab:abl-noise}. Performance remains stable up to 40\,mm noise, degrades moderately at 60\,mm, and drops more sharply beyond 80-100\,mm. This shows that \textit{SkelSplat} is resilient to realistic levels of initialization error, while reaching its limits under extreme perturbations.\looseness=-1

\begin{table}[tb]
\centering
\resizebox{\columnwidth}{!}{%
\begin{tabular}{lcccc}
\toprule
Absolute MPJPE, mm & 3 channels~\cite{kerbl20233d} & 1 channel & One-hot enc. & Impr. (\%) $\uparrow$\\ 
\midrule\midrule
\textbf{SkelSplat} (MeTRAbs~\cite{Sarandi_2023_WACV}) & 29.4 & 29.4 & \cellcolor{blue!8}\textbf{26.7} & \cellcolor{blue!8}+ 10.1 \\ 
\textbf{SkelSplat} (ResNet-152) & 21.9 & 21.9 & \cellcolor{blue!8}\textbf{20.3} & \cellcolor{blue!8}+ 7.9 \\
\bottomrule\bottomrule
\end{tabular}%
}
\caption{Ablation on joint encoding strategies for rendering.}
\label{tab:abl-encoding}
\end{table}

\begin{table}[tb]
\centering
\resizebox{0.7\columnwidth}{!}{%
\begin{tabular}{lccc}
\toprule
Absolute MPJPE, mm           & 1 view & 2 views    & 4 views       \\ \midrule \midrule
\textbf{SkelSplat} (MeTRAbs~\cite{Sarandi_2023_WACV}) & 30.9   &  \underline{28.2} & \textbf{26.7} \\ 
\textbf{SkelSplat} (ResNet-152) & 23.0   &  \underline{21.5} & \textbf{20.3} \\\bottomrule \bottomrule
\end{tabular}
}
\caption{Results on Human3.6M accumulating gradient across 1, 2 or 4 views.}
\label{tab:abl-accumulation}
\end{table}

\begin{table}[tb]
\centering
\resizebox{\columnwidth}{!}{%
\begin{tabular}{lcllll}
\toprule
Absolute MPJPE, mm & 4 views & 5 views & 6 views & 7 views & 8 views \\ \midrule \midrule
\textbf{SkelSplat} (MeTRAbs~\cite{Sarandi_2023_WACV})  & 20.9    &   17.7 &    \textbf{14.7} &    15.7     &   \underline{15.6}      \\
\bottomrule \bottomrule
\end{tabular}%
}
\caption{Impact of using different number of views for \textit{SkelSplat} optimization. Results on the CMU Panoptic Studio dataset.}
\label{tab:abl-views}
\end{table}

\begin{table}[tb]
\centering
\resizebox{0.8\columnwidth}{!}{%
\begin{tabular}{llccc}
\toprule
Init. method                                          & 2D poses   & Guess & SkelSplat & Impr. (\%) $\uparrow$\\ \midrule \midrule
\multirow{3}{*}{Triangulation}                        & CPN        & 30.6        & 27.4            & + 10.5     \\
                                                      & MeTRAbs~\cite{Sarandi_2023_WACV}    & 29.0        & \underline{26.7}            & + 7.9      \\
                                                      & ResNet-152 & 23.7        & \textbf{20.7}   & + 12.7     \\ \midrule
\multirow{3}{*}{3D Fusion~\cite{bragagnolo2025multi}} & CPN        & 40.2        & \underline{26.2}            & + 34.8     \\
                                                      & MeTRAbs~\cite{Sarandi_2023_WACV}    & 40.2        & 26.7            & + 33.6     \\
                                                      & ResNet-152 & 40.2        & \textbf{20.3}   & + 49.5     \\ \bottomrule \bottomrule
\end{tabular}%
}
\caption{Human3.6M results with different initialization strategies for 3D joint positions.}
\label{tab:abl-initial-guess}
\end{table}

\paragraph{Loss contribution to the final prediction} We conduct an ablation on the effect of the 3D structural symmetry loss applied to different subsets of body limbs, considering Human3.6M and its occluded variants. The baseline setting (Symm-1), used in our proposed \textit{SkelSplat}, applies the symmetry constraint to the lower arms and lower legs. We also evaluate extending the constraint to the upper arms and legs (Symm-2), and further to the torso limbs (Symm-3). 
On Human3.6M-Occ-2, Symm-1 yields the largest accuracy gain (+6.1\%). Symm-2 and Symm-3 provide marginal improvements with respect to it (+3.3\% and +3.7\%) but at higher cost, with Symm-3 increasing optimization time by +43.5\% over Symm-1. These results confirm Symm-1 as the best trade-off between performance and efficiency and is our proposed solution. More details are provided in the supplementary.\looseness=-1

\paragraph{Effect of rendering quality} We evaluate how rendering resolution affects pose accuracy by downscaling input images to 75\%, 50\%, and 25\% of their original size. Higher resolutions (100-75\%) yield the best performance with 20.3 and 20.4\,mm errors on Human3.6M, while accuracy degrades slightly at lower scales, with errors of 20.7 and 21.7\,mm at 50\% and 25\%. This demonstrates that \textit{SkelSplat} benefits from high-resolution inputs but remains robust to reduced rendering quality. Further details are provided in the supplementary.\looseness=-1

\paragraph{Impact of 2D covariance on pseudo ground truth} We also evaluate the effect of enlarging the covariance for frequently occluded joints, such as elbows, hands, and knees, on Human3.6-Occ-2 and Human3.6-Occ-3.
A moderate increase (1.25×) allows the model to better accommodate uncertainty in occluded regions, improving robustness. Larger covariances, however, reduce precision (+4.8\% error on average at 1.5× and nearly doubled error at 2×), as optimization becomes overly tolerant to noisy inputs. This highlights the need to balance flexibility and spatial accuracy for optimal performance. Extended ablation results are provided in the supplementary.\looseness=-1

\begin{table}[tb]
\centering
\resizebox{\columnwidth}{!}{%
\begin{tabular}{lccccccc}
\toprule
Noise $\sigma$, mm & 0 & 10 & 20 & 40 & 60 & 80 & 100 \\
\midrule\midrule
\textbf{SkelSplat} (MeTRAbs~\cite{Sarandi_2023_WACV}) & 26.7 & 27.8 & 27.9 & 30.6 & 41.5 & 63.4 & 84.6 \\
Variation (\%) & - & \textcolor[RGB]{0, 128, 0}{+ 4.1} & \textcolor[RGB]{0, 128, 0}{+ 4.5} & \textcolor{orange}{+ 14.6} & \textcolor{orange}{+ 55.4} & \textcolor{red}{+ 137.5} & \textcolor{red}{+ 216.8} \\                                       
\textbf{SkelSplat} (ResNet-152) & 20.3 & 21.5 & 22.0 & 25.4 & 38.3 & 61.9 & 83.6 \\
Variation (\%) & - & \textcolor[RGB]{0, 128, 0}{+ 5.9} & \textcolor[RGB]{0, 128, 0}{+ 8.4} & \textcolor{orange}{+ 25.1} & \textcolor{red}{+ 88.7} & \textcolor{red}{+ 204.9} & \textcolor{red}{+ 311.3} \\ 
\bottomrule\bottomrule
\end{tabular}%
}
\caption{Ablation on robustness to poor initialization, absolute MPJPE and relative variation under increasing Gaussian noise.}

\label{tab:abl-noise}
\end{table}


\subsection{Discussion}
Building on the success of Gaussian Splatting in dense 3D reconstruction~\cite{Wei_2025_CVPR, peng2023implicit, hu2024gauhuman, wen2024gomavatar}, where continuous optimization over a differentiable representation has proven to be more effective than direct geometric methods, we reinterpret multi-view pose estimation as a reconstruction problem. Unlike traditional Gaussian Splatting pipelines supervised by appearance, our supervision is driven by skeletal pose features. 
The presented experiments focus mainly on proving the accuracy of the skeleton represented by the means of the Gaussians. However, we believe covariance can potentially give useful information on the uncertainty of the prediction. For example, we measure the percentage of ground-truth 3D joints that lie within 1, 2, or 3 standard deviations from the corresponding Gaussian means on the Human3.6M dataset. We find that on average 98.0\% of joints fall within 3 sigmas, 91.6\% within 2 sigmas, and 57.2\% within 1 sigma. Lower coverage is typically observed for joints more prone to occlusions and self-occlusion, such as hands, elbows, and ankles. 
Further details are provided in the supplementary.
These results indicate that covariance can potentially include an estimate of the uncertainty of the joints.
However, more experiments are needed to prove the correlation. 
While this work focuses on robust pose estimation, inference speed may be a limiting factor for real-time applications. The current implementation requires about 3 seconds to estimate a pose from 4 views, as each joint is rendered in its own channel to allow independent optimization.
This motivates future work on more compact rendering schemes and alternative joint encodings that could significantly accelerate inference.
Extending \textit{SkelSplat} to multi-person scenarios by incorporating instance association across views to provide per-person 3D initialization, would further broaden its applicability in real-world environments.\looseness=-1

\section{Conclusions}

We presented \textit{SkelSplat}, a novel framework for multi-view 3D human pose estimation that departs from standard learning-based fusion strategies and instead leverages Gaussian Splatting for robust 3D human pose reconstruction. Unlike current methods, which rely heavily on training data and are tied to specific camera setups, pose distributions, or occlusion conditions, \textit{SkelSplat} is inherently flexible and generalizes effectively to novel environments without retraining or fine-tuning.
The experiments demonstrated that \textit{SkelSplat} achieves state-of-the-art performance across multiple benchmarks, showing strong robustness to occlusions and adaptability to different camera configurations, including out-of-distribution scenarios.
%
%
While our method offers accuracy and generalizability, it is not currently real-time as it operates by rendering a separate channel for each joint. 
Future work will explore more efficient formulations through compact joint encodings, while also investigating the combination of 3D human reconstruction with skeleton estimation, in particular using 3D reconstruction losses for 2D refinement to improve robustness. We further plan to extend \textit{SkelSplat} to multi-person scenarios, broadening its applicability in real-world settings.
{
    \small
    \bibliographystyle{ieeenat_fullname}
    \bibliography{main}
}


\renewcommand{\maketitlesupplementary}{%
  \clearpage 
  \twocolumn[%
    \begin{center}
      {\Large \bfseries \thetitle\par}
      \vspace{1.5em}
      {\Large \bfseries Supplementary Materials\par}
      \vspace{1.5em}
    \end{center}
  ]%
}
\maketitlesupplementary

\appendix
\section{Pseudo Ground Truth Generation}
This section provides additional details on how we generate the pseudo ground truth used during the optimization of \textit{SkelSplat}.
We begin by recalling that a human skeleton is defined as a set of joints $\mathrm{SK} = \{sk_0, \dots, sk_N\}$, each of which is represented by an anisotropic 3D Gaussian. This yields a corresponding set of Gaussians $\mathrm{GS} = \{gs_0, \dots, gs_N\}$, where each Gaussian $gs_j$ encodes the spatial uncertainty around joint $sk_j$ in 3D space.
The optimization process leverages 2D keypoint detections obtained from a pre-trained 2D human pose estimator. Specifically, for each camera view $i$ in a set of $M$ synchronized and calibrated views (i.e., $i \in \{1, \dots, M\}$), we extract 2D keypoint locations $\mathrm{SK^{2D}_i} = \{sk^{2D}_{i0}, \dots, sk^{2D}_{iN}\}$ corresponding to the projection of each joint in the image plane.
To supervise the optimization with view-dependent supervision, we generate a set of pseudo ground truth heatmaps $\{I_{ij}\}_{j=1}^{N}$ for each camera view. Each heatmap $I_{ij} \in \mathbb{R}^{H \times W}$ represents a soft target for joint $j$ in view $i$, and is constructed by rendering a 2D Gaussian $gs_{ij}^{2D}$ centered at the detected 2D location $sk_{ij}^{2D} \in \mathbb{R}^2$. 
The shape of this Gaussian is determined by a covariance matrix $\Sigma_{ij}^{2D} \in \mathbb{R}^{2 \times 2}$, which is obtained by projecting the original 3D covariance $\Sigma_j$ of joint $j$ into the 2D image plane of camera $i$ as follows:
\begin{equation}
\Sigma^{2D}_{ij} = J_i W_i \Sigma_j W^T_i J^T_i \,,
\end{equation}
where ${W}_i \in \mathbb{R}^{4 \times 4}$ is the camera extrinsic transformation (world-to-camera), and ${J}_i \in \mathbb{R}^{2 \times 3}$ is the Jacobian matrix of the perspective projection at the joint position in camera coordinates.
In detail, if $\mu_j \in \mathbb{R}^3$ is the 3D joint position, its homogeneous coordinate $\tilde{\mu}_j = [\mu_j^\top, 1]^\top$ is transformed into camera coordinates as:
\begin{equation}
\tilde{\mu}_{ij}^{\text{cam}} = {W}_i \tilde{\mu}_j = \begin{bmatrix} X_{ij} \\ Y_{ij} \\ Z_{ij} \\ 1 \end{bmatrix} \,.
\end{equation}
The Jacobian ${J}_i$ for the perspective projection with focal lengths $f_{x,i}, f_{y,i}$ is:
\begin{equation}
{J}_i = \begin{bmatrix}
\frac{f_{x,i}}{Z_{ij}} & 0 & -\frac{f_{x,i} X_{ij}}{Z_{ij}^2} \\
0 & \frac{f_{y,i}}{Z_{ij}} & -\frac{f_{y,i} Y_{ij}}{Z_{ij}^2}
\end{bmatrix}\,.
\end{equation}
A small constant $h$ is added to the covariance to prevent numerical issues.
To characterize the shape of the 2D covariance ellipse, we compute the eigenvalues $\lambda_1$ and $\lambda_2$ of $\Sigma^{2D}_{ij}$, which correspond to the principal axes variances:
\begin{align}
\det &= \Sigma^{2D}_{ij}(1,1) \cdot \Sigma^{2D}_{ij}(2,2) - \bigl(\Sigma^{2D}_{ij}(1,2)\bigr)^2 \,, \\
m &= \frac{\Sigma^{2D}_{ij}(1,1) + \Sigma^{2D}_{ij}(2,2)}{2} \,, \\
\lambda_1 &= m + \sqrt{\max(\epsilon, m^2 - \det)} \,, \\
\lambda_2 &= m - \sqrt{\max(\epsilon, m^2 - \det)} \,,
\end{align}
where $\epsilon$ is a small positive constant to ensure numerical stability, $(s, q)$ indicates the element with $s, q$ coordinates in the matrix.

\begin{figure*}[h!]
  \centering
  \includegraphics[width=1.7\columnwidth]{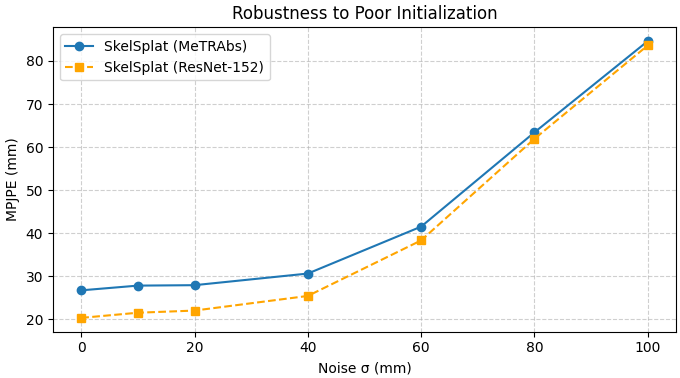}
  \caption{Ablation on robustness to poor initialization, adding Gaussian noise to triangulated joints.}
  \label{fig:abl-noise}
\end{figure*}

\section{Additional Details on Ablation Studies}
In this section, we present further details on the ablation studies discussed in the main paper. Specifically, we include expanded analyses of noise robustness (as shown in~\cref{fig:abl-noise}), loss component contributions (see ~\cref{tab:abl-symmetry}), covariance scaling behaviors (see ~\cref{tab:abl-covariance}), and the effects of rendering resolution (see ~\cref{tab:abl-resolution}).

\paragraph{Robustness to noisy initialization} On Human3.6M~\cite{ionescu2013human3}, we perturb triangulation of MeTRAbs~\cite{Sarandi_2023_WACV} and ResNet-152 poses with Gaussian noise of increasing standard deviation applied independently to each joint. We consider values for $\sigma$ equal to 10, 20, 40, 60, 80 and 100\,mm. From~\cref{fig:abl-noise} we can observe how performance remains stable up to 40\,mm noise and starts to degrade at 60\,mm. With strong noise (80-100\,mm) accuracy drops more sharply.

\paragraph{Contribution of 3D loss to optimization} Results in~\cref{tab:abl-symmetry} show results for \textit{SkelSplat} using three variants of 3D symmetry loss during optimization: Symm-1 applies the symmetry constraint to the lower arms and lower legs, Symm-2 extends the constraint to the upper arms and legs and Symm-3 further adds constraints from the hip joints to the root and from the shoulders to the neck. ~\cref{tab:abl-symmetry} illustrates accuracy on Human3.6M~\cite{ionescu2013human3}, Human3.6M-Occ-2 and Human3.6M-Occ-3~\cite{bragagnolo2025multi}.

\paragraph{Effect of rendering resolution} We assess the impact of rendering resolution on pose accuracy by reducing the input images to 75\%, 50\%, and 25\% of their original resolution. ~\cref{tab:abl-resolution} reports full results for \textit{SkelSplat} using 2D detections from MeTRAbs and ResNet-152.

\paragraph{Impact of covariance scaling for 2D pseudo ground truth generation} We evaluate the effect of enlarging the covariance for frequently occluded joints, such as elbows, hands, and knees, on Human3.6-Occ-2 and Human3.6-Occ-3. 
\cref{tab:abl-covariance} reports results using scaling factors of 1.25, 1.5, and 2 applied to the default covariance of these joints. 
Overly large covariances decrease reconstruction accuracy, since the optimization becomes too tolerant of noisy or imprecise 2D inputs.

\begin{table}[tb]
\centering
\resizebox{0.85\columnwidth}{!}{%
\begin{tabular}{lcccc}
\toprule
Image resolution scale (\%) & 100 & 75 & 50 & 25 \\
\midrule\midrule
SkelSplat (MeTRAbs~\cite{Sarandi_2023_WACV}) & \textbf{26.7} & \textbf{26.7} & \underline{26.8} & 27.5 \\
SkelSplat (ResNet-152) & \textbf{20.3} & \underline{20.4} & 20.7 & 21.7 \\
\bottomrule\bottomrule
\end{tabular}%
}
\caption{Resolution ablation. We downscale inputs to a percentage of the original size before rendering; higher resolution yields slightly lower MPJPE (mm).}
\label{tab:abl-resolution}
\end{table}

\begin{table}[tb]
\centering
\resizebox{\columnwidth}{!}{%
\begin{tabular}{llcccc}
\toprule
Absolute MPJPE, mm               &            &                &                 &                &                \\ \midrule \midrule
\multirow{2}{*}{Human3.6M}       & MeTRAbs~\cite{Sarandi_2023_WACV}    & 27.0           & \textbf{26.7}   & 26.9           & \underline{26.8} \\
                                 & ResNet-152 & 20.6           & \textbf{20.3}            & \underline{20.4}          & \underline{20.4}           \\ \midrule
\multirow{2}{*}{Human3.6M-Occ-2} & MeTRAbs~\cite{Sarandi_2023_WACV}    & 30.0           & 29.6            & \underline{29.5} & \textbf{29.4}  \\
                                 & ResNet-152 & 26.2           & 24.6            & \underline{23.8}           & \textbf{23.7}           \\ \midrule
\multirow{2}{*}{Human3.6M-Occ-3} & MeTRAbs~\cite{Sarandi_2023_WACV}    & 31.4           & \underline{31.1}  & \textbf{31.0}  & \textbf{31.0}  \\
                                 & ResNet-152 & 27.2           & 27.0            & \textbf{26.0}          & \underline{26.1}           \\ \midrule \midrule
3D Symm-1                        &            & -              & \checkmark      & \checkmark     & \checkmark     \\
3D Symm-2                        &            & -              & -               & \checkmark     & \checkmark     \\
3D Symm-3                        &            & -              & -               & -              & \checkmark     \\ \hline
Sec/iter                         &            & \textbf{0.028} & \underline{0.039} & 0.045          & 0.056          \\ \bottomrule \bottomrule
\end{tabular}%
}
\caption{Ablation study on different 3D symmetry loss contributions on the Human3.6M dataset and its occluded versions.}
\label{tab:abl-symmetry}
\end{table}

\begin{table}[t]
\centering
\resizebox{0.8\columnwidth}{!}{%
\begin{tabular}{llccc}
\toprule
Scaling factor               &            & 1.25                              & 1.5                      & 2.0                      \\ \midrule \midrule
\multirow{2}{*}{H3.6M-Occ-2} & MeTRAbs~\cite{Sarandi_2023_WACV}    & \textbf{29.6}                     & \underline{31.0}           & 60.5                     \\
                             & ResNet-152 & \textbf{24.6} & \underline{26.0} & 54.3 \\ \midrule
\multirow{2}{*}{H3.6M-Occ-3} & MeTRAbs~\cite{Sarandi_2023_WACV}    & \textbf{31.1}                     & \underline{32.6}           & 61.9                     \\
                             & ResNet-152 & \textbf{27.0}                     & \underline{28.9}                     & 59.0                     \\ \bottomrule \bottomrule
\end{tabular}%
}
\caption{Ablation study on different methods to initialize 3D joint positions, absolute MPJPE (mm).}
\label{tab:abl-covariance}
\end{table}

\section{Joint Covariance for Confidence Estimation}

\begin{figure*}[tb]
  \centering
    \includegraphics[width=1.85\columnwidth]{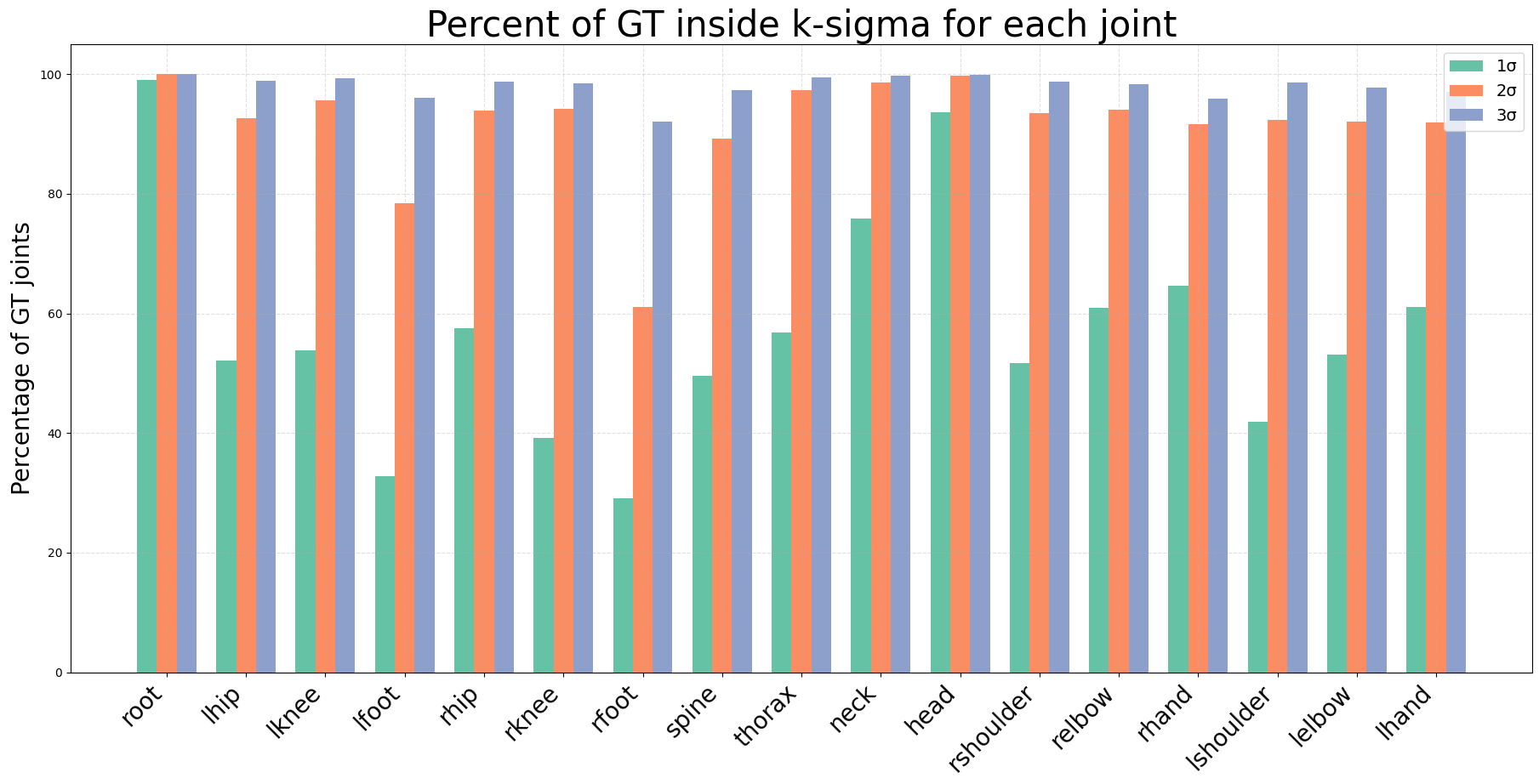}
    \caption{Joint-wise coverage rates across different sigma thresholds for the Human3.6M dataset.}
    \label{fig:sigma_h36m}
\end{figure*}

\begin{figure*}[tb]
    \centering
    \includegraphics[width=1.85\columnwidth]{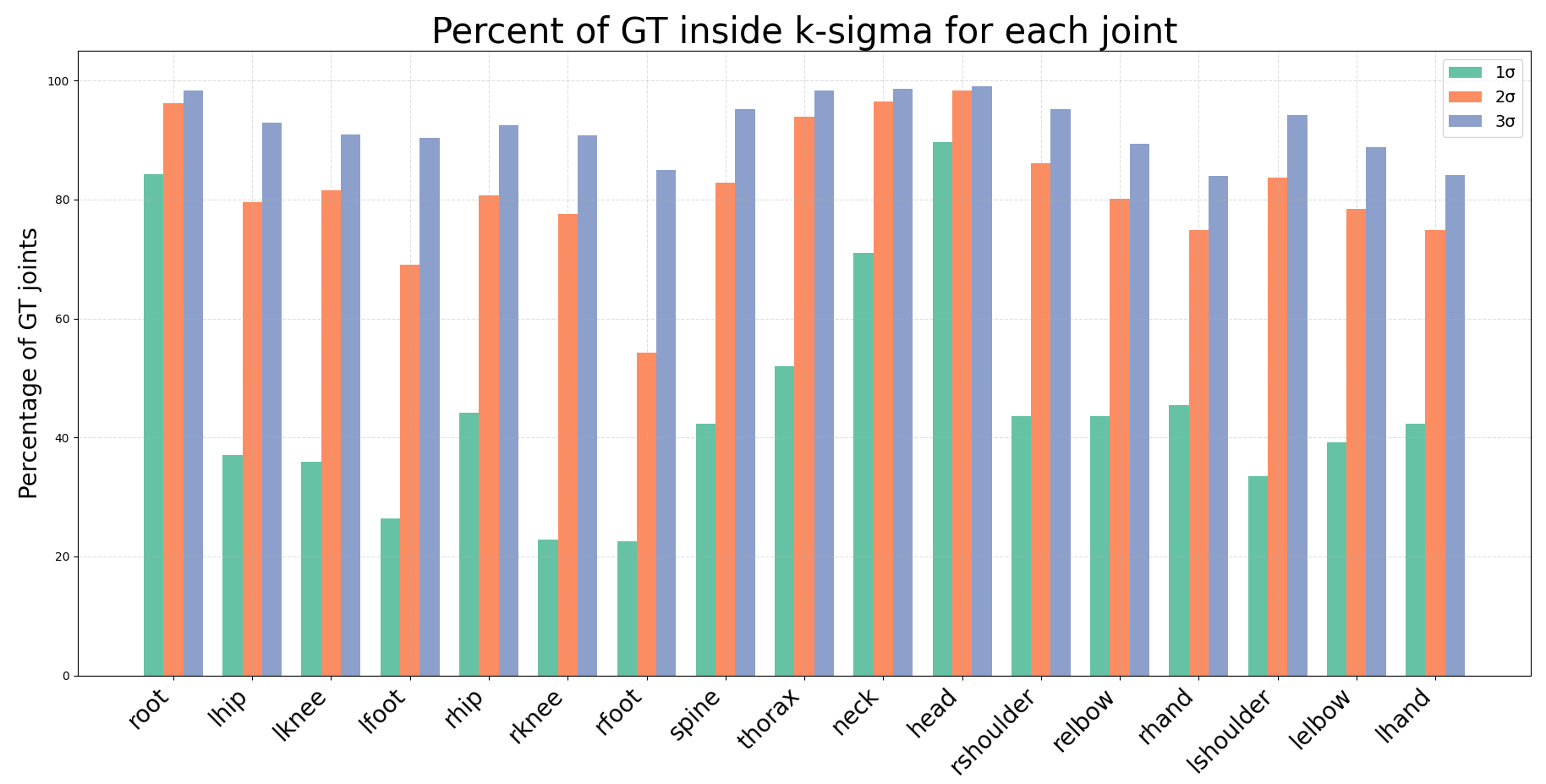}
    \caption{Joint-wise coverage rates across different sigma thresholds for the Human3.6M-Occ-3 dataset.}
    \label{fig:sigma_occ}
\end{figure*}

While our experiments primarily evaluate the 3D means of the Gaussians as joint predictions, the associated covariances also encode potentially useful information about prediction uncertainty.
To assess this, we compute the percentage of ground-truth joints that fall within 1, 2, or 3 standard deviations of the predicted Gaussian means on the Human3.6M dataset and on its occluded version Human3.6M-Occ-3. Here, in~\cref{fig:sigma_h36m} and~\cref{fig:sigma_occ} we include results for joint-wise coverage in both occluded and non-occluded settings. Notably, in both cases, joints that are often occluded or self-occluded, such as hands, elbows, and ankles, tend to exhibit lower coverage.

\section{Supplementary Visualizations}
\begin{figure*}[hbt]
  \centering
  \includegraphics[width=2\columnwidth]{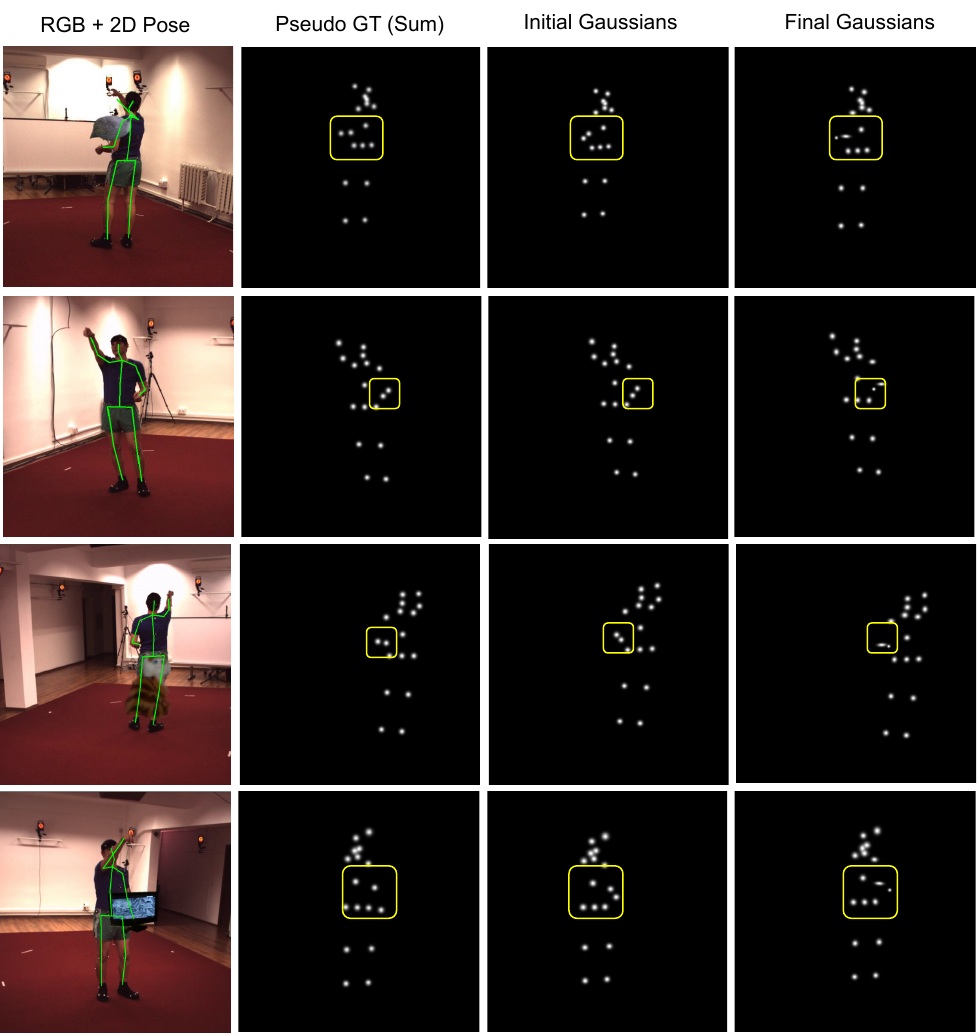}
  \caption{Visualization of pseudo ground-truth supervision and joint refinement. For 4 camera views, we show (left) the aggregated pseudo ground-truth heatmap obtained by summing the 2D Gaussian heatmaps of all joints, and (right) a 3D visualization of the joint Gaussians before and after optimization.}
  \label{fig:visualization}
\end{figure*}

In~\cref{fig:visualization}, we provide a visualization for a scene with four camera views from Human3.6M-Occ-3. We show the aggregated pseudo ground truth heatmap, obtained by summing the per-joint 2D heatmaps across all joints, together with a comparison between the initial and the final (optimized) 3D joint Gaussians. This highlights how the optimization step progressively refines the 3D pose, leading to improved alignment with the set of multi-view 2D detections and producing a more coherent reconstruction.

\begin{figure*}[hbt]
  \centering
  \includegraphics[width=2\columnwidth]{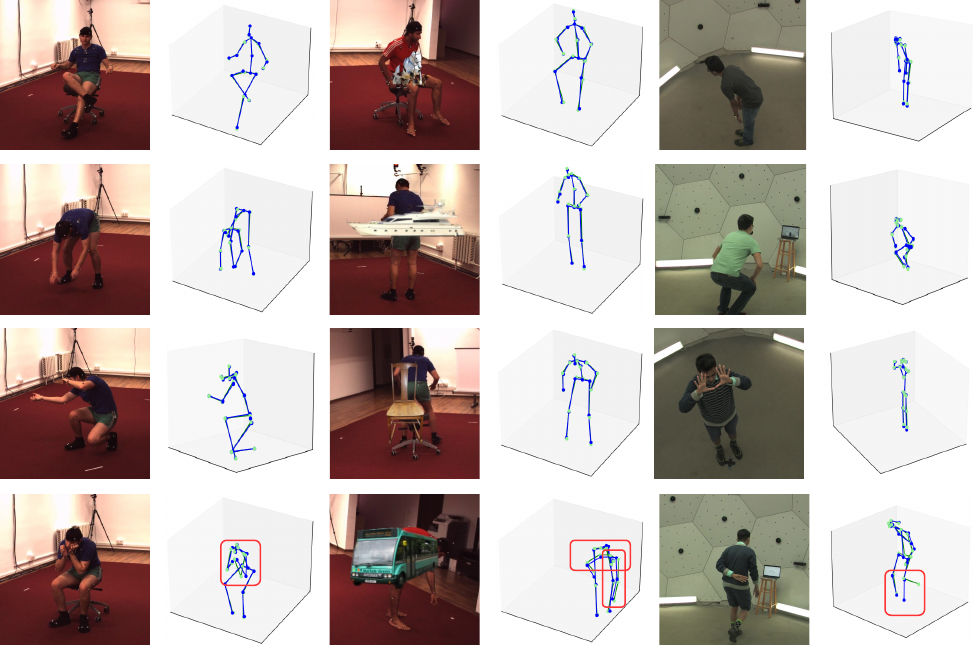}
  \caption{Qualitative results on Human3.6M (left), Human3.6M-Occ-3 (middle), and CMU Panoptic (right). For Human3.6M-Occ-3 we show one of the three occluded views. The last row shows some failure cases.}
  \label{fig:qualitative}
\end{figure*}

In~\cref{fig:qualitative}, we report additional qualitative results on Human3.6M, Human3.6M-Occ-3, and the CMU Panoptic Studio datasets. For Human3.6M-Occ-3, we show only one of the occluded camera views. final row presents representative failure cases, in which our method struggles due to factors such as extreme occlusion or complex limb configurations, for which inconsistent multi-view evidence is common.

\clearpage
\end{document}